\newcounter{treecount}
\newcounter{branchcount}
\newsavebox{\parentbox}
\newsavebox{\treebox}
\newsavebox{\treeboxone}
\newsavebox{\treeboxtwo}
\newsavebox{\treeboxthree}
\newsavebox{\treeboxfour}
\newsavebox{\treeboxfive}
\newsavebox{\treeboxsix}
\newsavebox{\treeboxseven}
\newsavebox{\treeboxeight}
\newsavebox{\treeboxnine}
\newsavebox{\treeboxten}
\newsavebox{\treeboxeleven}
\newsavebox{\treeboxtwelve}
\newsavebox{\treeboxthirteen}
\newsavebox{\treeboxfourteen}
\newsavebox{\treeboxfifteen}
\newsavebox{\treeboxsixteen}
\newsavebox{\treeboxseventeen}
\newsavebox{\treeboxeighteen}
\newsavebox{\treeboxnineteen}
\newsavebox{\treeboxtwenty}
\newlength{\treeoffsetone}
\newlength{\treeoffsettwo}
\newlength{\treeoffsetthree}
\newlength{\treeoffsetfour}
\newlength{\treeoffsetfive}
\newlength{\treeoffsetsix}
\newlength{\treeoffsetseven}
\newlength{\treeoffseteight}
\newlength{\treeoffsetnine}
\newlength{\treeoffsetten}
\newlength{\treeoffseteleven}
\newlength{\treeoffsettwelve}
\newlength{\treeoffsetthirteen}
\newlength{\treeoffsetfourteen}
\newlength{\treeoffsetfifteen}
\newlength{\treeoffsetsixteen}
\newlength{\treeoffsetseventeen}
\newlength{\treeoffseteighteen}
\newlength{\treeoffsetnineteen}
\newlength{\treeoffsettwenty}

\newlength{\treeshiftone}
\newlength{\treeshifttwo}
\newlength{\treeshiftthree}
\newlength{\treeshiftfour}
\newlength{\treeshiftfive}
\newlength{\treeshiftsix}
\newlength{\treeshiftseven}
\newlength{\treeshifteight}
\newlength{\treeshiftnine}
\newlength{\treeshiftten}
\newlength{\treeshifteleven}
\newlength{\treeshifttwelve}
\newlength{\treeshiftthirteen}
\newlength{\treeshiftfourteen}
\newlength{\treeshiftfifteen}
\newlength{\treeshiftsixteen}
\newlength{\treeshiftseventeen}
\newlength{\treeshifteighteen}
\newlength{\treeshiftnineteen}
\newlength{\treeshifttwenty}
\newlength{\treewidthone}
\newlength{\treewidthtwo}
\newlength{\treewidththree}
\newlength{\treewidthfour}
\newlength{\treewidthfive}
\newlength{\treewidthsix}
\newlength{\treewidthseven}
\newlength{\treewidtheight}
\newlength{\treewidthnine}
\newlength{\treewidthten}
\newlength{\treewidtheleven}
\newlength{\treewidthtwelve}
\newlength{\treewidththirteen}
\newlength{\treewidthfourteen}
\newlength{\treewidthfifteen}
\newlength{\treewidthsixteen}
\newlength{\treewidthseventeen}
\newlength{\treewidtheighteen}
\newlength{\treewidthnineteen}
\newlength{\treewidthtwenty}
\newlength{\daughteroffsetone}
\newlength{\daughteroffsettwo}
\newlength{\daughteroffsetthree}
\newlength{\daughteroffsetfour}
\newlength{\branchwidthone}
\newlength{\branchwidthtwo}
\newlength{\branchwidththree}
\newlength{\branchwidthfour}
\newlength{\parentoffset}
\newlength{\treeoffset}
\newlength{\daughteroffset}
\newlength{\branchwidth}
\newlength{\parentwidth}
\newlength{\treewidth}
\newcommand{\ontop}[1]{\begin{tabular}{c}#1\end{tabular}}
\newcommand{\poptree}{%
\ifnum\value{treecount}=0\typeout{QobiTeX warning---Tree stack underflow}\fi%
\addtocounter{treecount}{-1}%
\setlength{\treeoffsettwo}{\treeoffsetthree}%
\setlength{\treeoffsetthree}{\treeoffsetfour}%
\setlength{\treeoffsetfour}{\treeoffsetfive}%
\setlength{\treeoffsetfive}{\treeoffsetsix}%
\setlength{\treeoffsetsix}{\treeoffsetseven}%
\setlength{\treeoffsetseven}{\treeoffseteight}%
\setlength{\treeoffseteight}{\treeoffsetnine}%
\setlength{\treeoffsetnine}{\treeoffsetten}%
\setlength{\treeoffsetten}{\treeoffseteleven}%
\setlength{\treeoffseteleven}{\treeoffsettwelve}%
\setlength{\treeoffsettwelve}{\treeoffsetthirteen}%
\setlength{\treeoffsetthirteen}{\treeoffsetfourteen}%
\setlength{\treeoffsetfourteen}{\treeoffsetfifteen}%
\setlength{\treeoffsetfifteen}{\treeoffsetsixteen}%
\setlength{\treeoffsetsixteen}{\treeoffsetseventeen}%
\setlength{\treeoffsetseventeen}{\treeoffseteighteen}%
\setlength{\treeoffseteighteen}{\treeoffsetnineteen}%
\setlength{\treeoffsetnineteen}{\treeoffsettwenty}%
\setlength{\treeshifttwo}{\treeshiftthree}%
\setlength{\treeshiftthree}{\treeshiftfour}%
\setlength{\treeshiftfour}{\treeshiftfive}%
\setlength{\treeshiftfive}{\treeshiftsix}%
\setlength{\treeshiftsix}{\treeshiftseven}%
\setlength{\treeshiftseven}{\treeshifteight}%
\setlength{\treeshifteight}{\treeshiftnine}%
\setlength{\treeshiftnine}{\treeshiftten}%
\setlength{\treeshiftten}{\treeshifteleven}%
\setlength{\treeshifteleven}{\treeshifttwelve}%
\setlength{\treeshifttwelve}{\treeshiftthirteen}%
\setlength{\treeshiftthirteen}{\treeshiftfourteen}%
\setlength{\treeshiftfourteen}{\treeshiftfifteen}%
\setlength{\treeshiftfifteen}{\treeshiftsixteen}%
\setlength{\treeshiftsixteen}{\treeshiftseventeen}%
\setlength{\treeshiftseventeen}{\treeshifteighteen}%
\setlength{\treeshifteighteen}{\treeshiftnineteen}%
\setlength{\treeshiftnineteen}{\treeshifttwenty}%
\setlength{\treewidthtwo}{\treewidththree}%
\setlength{\treewidththree}{\treewidthfour}%
\setlength{\treewidthfour}{\treewidthfive}%
\setlength{\treewidthfive}{\treewidthsix}%
\setlength{\treewidthsix}{\treewidthseven}%
\setlength{\treewidthseven}{\treewidtheight}%
\setlength{\treewidtheight}{\treewidthnine}%
\setlength{\treewidthnine}{\treewidthten}%
\setlength{\treewidthten}{\treewidtheleven}%
\setlength{\treewidtheleven}{\treewidthtwelve}%
\setlength{\treewidthtwelve}{\treewidththirteen}%
\setlength{\treewidththirteen}{\treewidthfourteen}%
\setlength{\treewidthfourteen}{\treewidthfifteen}%
\setlength{\treewidthfifteen}{\treewidthsixteen}%
\setlength{\treewidthsixteen}{\treewidthseventeen}%
\setlength{\treewidthseventeen}{\treewidtheighteen}%
\setlength{\treewidtheighteen}{\treewidthnineteen}%
\setlength{\treewidthnineteen}{\treewidthtwenty}%
\sbox{\treeboxtwo}{\usebox{\treeboxthree}}%
\sbox{\treeboxthree}{\usebox{\treeboxfour}}%
\sbox{\treeboxfour}{\usebox{\treeboxfive}}%
\sbox{\treeboxfive}{\usebox{\treeboxsix}}%
\sbox{\treeboxsix}{\usebox{\treeboxseven}}%
\sbox{\treeboxseven}{\usebox{\treeboxeight}}%
\sbox{\treeboxeight}{\usebox{\treeboxnine}}%
\sbox{\treeboxnine}{\usebox{\treeboxten}}%
\sbox{\treeboxten}{\usebox{\treeboxeleven}}%
\sbox{\treeboxeleven}{\usebox{\treeboxtwelve}}%
\sbox{\treeboxtwelve}{\usebox{\treeboxthirteen}}%
\sbox{\treeboxthirteen}{\usebox{\treeboxfourteen}}%
\sbox{\treeboxfourteen}{\usebox{\treeboxfifteen}}%
\sbox{\treeboxfifteen}{\usebox{\treeboxsixteen}}%
\sbox{\treeboxsixteen}{\usebox{\treeboxseventeen}}%
\sbox{\treeboxseventeen}{\usebox{\treeboxeighteen}}%
\sbox{\treeboxeighteen}{\usebox{\treeboxnineteen}}%
\sbox{\treeboxnineteen}{\usebox{\treeboxtwenty}}}
\newcommand{\leaf}[1]{%
\ifnum\value{treecount}=20\typeout{QobiTeX warning---Tree stack overflow}\fi%
\addtocounter{treecount}{1}%
\sbox{\treeboxtwenty}{\usebox{\treeboxnineteen}}%
\sbox{\treeboxnineteen}{\usebox{\treeboxeighteen}}%
\sbox{\treeboxeighteen}{\usebox{\treeboxseventeen}}%
\sbox{\treeboxseventeen}{\usebox{\treeboxsixteen}}%
\sbox{\treeboxsixteen}{\usebox{\treeboxfifteen}}%
\sbox{\treeboxfifteen}{\usebox{\treeboxfourteen}}%
\sbox{\treeboxfourteen}{\usebox{\treeboxthirteen}}%
\sbox{\treeboxthirteen}{\usebox{\treeboxtwelve}}%
\sbox{\treeboxtwelve}{\usebox{\treeboxeleven}}%
\sbox{\treeboxeleven}{\usebox{\treeboxten}}%
\sbox{\treeboxten}{\usebox{\treeboxnine}}%
\sbox{\treeboxnine}{\usebox{\treeboxeight}}%
\sbox{\treeboxeight}{\usebox{\treeboxseven}}%
\sbox{\treeboxseven}{\usebox{\treeboxsix}}%
\sbox{\treeboxsix}{\usebox{\treeboxfive}}%
\sbox{\treeboxfive}{\usebox{\treeboxfour}}%
\sbox{\treeboxfour}{\usebox{\treeboxthree}}%
\sbox{\treeboxthree}{\usebox{\treeboxtwo}}%
\sbox{\treeboxtwo}{\usebox{\treeboxone}}%
\sbox{\treeboxone}{\ontop{#1}}%
\sbox{\treeboxone}{\raisebox{-\ht\treeboxone}{\usebox{\treeboxone}}}%
\setlength{\treeoffsettwenty}{\treeoffsetnineteen}%
\setlength{\treeoffsetnineteen}{\treeoffseteighteen}%
\setlength{\treeoffseteighteen}{\treeoffsetseventeen}%
\setlength{\treeoffsetseventeen}{\treeoffsetsixteen}%
\setlength{\treeoffsetsixteen}{\treeoffsetfifteen}%
\setlength{\treeoffsetfifteen}{\treeoffsetfourteen}%
\setlength{\treeoffsetfourteen}{\treeoffsetthirteen}%
\setlength{\treeoffsetthirteen}{\treeoffsettwelve}%
\setlength{\treeoffsettwelve}{\treeoffseteleven}%
\setlength{\treeoffseteleven}{\treeoffsetten}%
\setlength{\treeoffsetten}{\treeoffsetnine}%
\setlength{\treeoffsetnine}{\treeoffseteight}%
\setlength{\treeoffseteight}{\treeoffsetseven}%
\setlength{\treeoffsetseven}{\treeoffsetsix}%
\setlength{\treeoffsetsix}{\treeoffsetfive}%
\setlength{\treeoffsetfive}{\treeoffsetfour}%
\setlength{\treeoffsetfour}{\treeoffsetthree}%
\setlength{\treeoffsetthree}{\treeoffsettwo}%
\setlength{\treeoffsettwo}{\treeoffsetone}%
\setlength{\treeoffsetone}{0.5\wd\treeboxone}%
\setlength{\treeshifttwenty}{\treeshiftnineteen}%
\setlength{\treeshiftnineteen}{\treeshifteighteen}%
\setlength{\treeshifteighteen}{\treeshiftseventeen}%
\setlength{\treeshiftseventeen}{\treeshiftsixteen}%
\setlength{\treeshiftsixteen}{\treeshiftfifteen}%
\setlength{\treeshiftfifteen}{\treeshiftfourteen}%
\setlength{\treeshiftfourteen}{\treeshiftthirteen}%
\setlength{\treeshiftthirteen}{\treeshifttwelve}%
\setlength{\treeshifttwelve}{\treeshifteleven}%
\setlength{\treeshifteleven}{\treeshiftten}%
\setlength{\treeshiftten}{\treeshiftnine}%
\setlength{\treeshiftnine}{\treeshifteight}%
\setlength{\treeshifteight}{\treeshiftseven}%
\setlength{\treeshiftseven}{\treeshiftsix}%
\setlength{\treeshiftsix}{\treeshiftfive}%
\setlength{\treeshiftfive}{\treeshiftfour}%
\setlength{\treeshiftfour}{\treeshiftthree}%
\setlength{\treeshiftthree}{\treeshifttwo}%
\setlength{\treeshifttwo}{\treeshiftone}%
\setlength{\treeshiftone}{0pt}%
\setlength{\treewidthtwenty}{\treewidthnineteen}%
\setlength{\treewidthnineteen}{\treewidtheighteen}%
\setlength{\treewidtheighteen}{\treewidthseventeen}%
\setlength{\treewidthseventeen}{\treewidthsixteen}%
\setlength{\treewidthsixteen}{\treewidthfifteen}%
\setlength{\treewidthfifteen}{\treewidthfourteen}%
\setlength{\treewidthfourteen}{\treewidththirteen}%
\setlength{\treewidththirteen}{\treewidthtwelve}%
\setlength{\treewidthtwelve}{\treewidtheleven}%
\setlength{\treewidtheleven}{\treewidthten}%
\setlength{\treewidthten}{\treewidthnine}%
\setlength{\treewidthnine}{\treewidtheight}%
\setlength{\treewidtheight}{\treewidthseven}%
\setlength{\treewidthseven}{\treewidthsix}%
\setlength{\treewidthsix}{\treewidthfive}%
\setlength{\treewidthfive}{\treewidthfour}%
\setlength{\treewidthfour}{\treewidththree}%
\setlength{\treewidththree}{\treewidthtwo}%
\setlength{\treewidthtwo}{\treewidthone}%
\setlength{\treewidthone}{\wd\treeboxone}}
\newcommand{\branch}[2]{%
\setcounter{branchcount}{#1}%
\ifnum\value{branchcount}=1\sbox{\parentbox}{\ontop{#2}}%
\setlength{\parentoffset}{\treeoffsetone}%
\addtolength{\parentoffset}{-0.5\wd\parentbox}%
\setlength{\daughteroffset}{0in}%
\ifdim\parentoffset<0in%
\setlength{\daughteroffset}{-\parentoffset}%
\setlength{\parentoffset}{0in}\fi%
\setlength{\parentwidth}{\parentoffset}%
\addtolength{\parentwidth}{\wd\parentbox}%
\setlength{\treeoffset}{\daughteroffset}%
\addtolength{\treeoffset}{\treeoffsetone}%
\setlength{\treewidth}{\wd\treeboxone}%
\addtolength{\treewidth}{\daughteroffset}%
\ifdim\treewidth<\parentwidth\setlength{\treewidth}{\parentwidth}\fi%
\sbox{\treebox}{\begin{minipage}{\treewidth}%
\begin{flushleft}%
\hspace*{\parentoffset}\usebox{\parentbox}\\
{\setlength{\unitlength}{2ex}%
\hspace*{\treeoffset}\begin{picture}(0,1)%
\put(0,0){\line(0,1){1}}%
\end{picture}}\\
\vspace{-\baselineskip}
\hspace*{\daughteroffset}%
\raisebox{-\ht\treeboxone}{\usebox{\treeboxone}}%
\end{flushleft}%
\end{minipage}}%
\setlength{\treeoffsetone}{\parentoffset}%
\addtolength{\treeoffsetone}{0.5\wd\parentbox}%
\setlength{\treeshiftone}{0pt}%
\setlength{\treewidthone}{\treewidth}%
\sbox{\treeboxone}{\usebox{\treebox}}%
\else\ifnum\value{branchcount}=2\sbox{\parentbox}{\ontop{#2}}%
\setlength{\branchwidthone}{\treewidthtwo}%
\addtolength{\branchwidthone}{\treeoffsetone}%
\addtolength{\branchwidthone}{-\treeshiftone}%
\addtolength{\branchwidthone}{-\treeoffsettwo}%
\setlength{\branchwidth}{\branchwidthone}%
\setlength{\daughteroffsetone}{\branchwidth}%
\addtolength{\daughteroffsetone}{-\branchwidthone}%
\addtolength{\daughteroffsetone}{-\treeshiftone}%
\setlength{\parentoffset}{-0.5\wd\parentbox}%
\addtolength{\parentoffset}{\treeoffsettwo}%
\addtolength{\parentoffset}{0.5\branchwidth}%
\setlength{\daughteroffset}{0in}%
\ifdim\parentoffset<0in%
\setlength{\daughteroffset}{-\parentoffset}%
\setlength{\parentoffset}{0in}\fi%
\setlength{\parentwidth}{\parentoffset}%
\addtolength{\parentwidth}{\wd\parentbox}%
\setlength{\treeoffset}{\daughteroffset}%
\addtolength{\treeoffset}{\treeoffsettwo}%
\setlength{\treewidth}{\wd\treeboxone}%
\addtolength{\treewidth}{\daughteroffsetone}%
\addtolength{\treewidth}{\treewidthtwo}%
\addtolength{\treewidth}{\daughteroffset}%
\ifdim\treewidth<\parentwidth\setlength{\treewidth}{\parentwidth}\fi%
\sbox{\treebox}{\begin{minipage}{\treewidth}%
\begin{flushleft}%
\hspace*{\parentoffset}\usebox{\parentbox}\\
{\setlength{\unitlength}{0.5\branchwidth}%
\hspace*{\treeoffset}\begin{picture}(2,0.5)%
\put(0,0){\line(2,1){1}}%
\put(2,0){\line(-2,1){1}}%
\end{picture}}\\
\vspace{-\baselineskip}
\hspace*{\daughteroffset}%
\makebox[\treewidthtwo][l]%
{\raisebox{-\ht\treeboxtwo}{\usebox{\treeboxtwo}}}%
\hspace*{\daughteroffsetone}%
\raisebox{-\ht\treeboxone}{\usebox{\treeboxone}}%
\end{flushleft}%
\end{minipage}}%
\setlength{\treeoffsetone}{\parentoffset}%
\addtolength{\treeoffsetone}{0.5\wd\parentbox}%
\setlength{\treeshiftone}{0pt}%
\setlength{\treewidthone}{\treewidth}%
\sbox{\treeboxone}{\usebox{\treebox}}\poptree%
\else\ifnum\value{branchcount}=3\sbox{\parentbox}{\ontop{#2}}%
\setlength{\branchwidthone}{\treewidthtwo}%
\addtolength{\branchwidthone}{\treeoffsetone}%
\addtolength{\branchwidthone}{-\treeshiftone}%
\addtolength{\branchwidthone}{-\treeoffsettwo}%
\setlength{\branchwidthtwo}{\treewidththree}%
\addtolength{\branchwidthtwo}{\treeoffsettwo}%
\addtolength{\branchwidthtwo}{-\treeshifttwo}%
\addtolength{\branchwidthtwo}{-\treeoffsetthree}%
\setlength{\branchwidth}{\branchwidthone}%
\ifdim\branchwidthtwo>\branchwidth%
\setlength{\branchwidth}{\branchwidthtwo}\fi%
\setlength{\daughteroffsetone}{\branchwidth}%
\addtolength{\daughteroffsetone}{-\branchwidthone}%
\addtolength{\daughteroffsetone}{-\treeshiftone}%
\setlength{\daughteroffsettwo}{\branchwidth}%
\addtolength{\daughteroffsettwo}{-\branchwidthtwo}%
\addtolength{\daughteroffsettwo}{-\treeshifttwo}%
\setlength{\parentoffset}{-0.5\wd\parentbox}%
\addtolength{\parentoffset}{\treeoffsetthree}%
\addtolength{\parentoffset}{\branchwidth}%
\setlength{\daughteroffset}{0in}%
\ifdim\parentoffset<0in%
\setlength{\daughteroffset}{-\parentoffset}%
\setlength{\parentoffset}{0in}\fi%
\setlength{\parentwidth}{\parentoffset}%
\addtolength{\parentwidth}{\wd\parentbox}%
\setlength{\treeoffset}{\daughteroffset}%
\addtolength{\treeoffset}{\treeoffsetthree}%
\setlength{\treewidth}{\wd\treeboxone}%
\addtolength{\treewidth}{\daughteroffsetone}%
\addtolength{\treewidth}{\treewidthtwo}%
\addtolength{\treewidth}{\daughteroffsettwo}%
\addtolength{\treewidth}{\treewidththree}%
\addtolength{\treewidth}{\daughteroffset}%
\ifdim\treewidth<\parentwidth\setlength{\treewidth}{\parentwidth}\fi%
\sbox{\treebox}{\begin{minipage}{\treewidth}%
\begin{flushleft}%
\hspace*{\parentoffset}\usebox{\parentbox}\\
{\setlength{\unitlength}{0.5\branchwidth}%
\hspace*{\treeoffset}\begin{picture}(4,1)%
\put(0,0){\line(2,1){2}}%
\put(2,0){\line(0,1){1}}%
\put(4,0){\line(-2,1){2}}%
\end{picture}}\\
\vspace{-\baselineskip}
\hspace*{\daughteroffset}%
\makebox[\treewidththree][l]%
{\raisebox{-\ht\treeboxthree}{\usebox{\treeboxthree}}}%
\hspace*{\daughteroffsettwo}%
\makebox[\treewidthtwo][l]%
{\raisebox{-\ht\treeboxtwo}{\usebox{\treeboxtwo}}}%
\hspace*{\daughteroffsetone}%
\raisebox{-\ht\treeboxone}{\usebox{\treeboxone}}%
\end{flushleft}%
\end{minipage}}%
\setlength{\treeoffsetone}{\parentoffset}%
\addtolength{\treeoffsetone}{0.5\wd\parentbox}%
\setlength{\treeshiftone}{0pt}%
\setlength{\treewidthone}{\treewidth}%
\sbox{\treeboxone}{\usebox{\treebox}}\poptree\poptree%
\else\ifnum\value{branchcount}=4\sbox{\parentbox}{\ontop{#2}}%
\setlength{\branchwidthone}{\treewidthtwo}%
\addtolength{\branchwidthone}{\treeoffsetone}%
\addtolength{\branchwidthone}{-\treeshiftone}%
\addtolength{\branchwidthone}{-\treeoffsettwo}%
\setlength{\branchwidthtwo}{\treewidththree}%
\addtolength{\branchwidthtwo}{\treeoffsettwo}%
\addtolength{\branchwidthtwo}{-\treeshifttwo}%
\addtolength{\branchwidthtwo}{-\treeoffsetthree}%
\setlength{\branchwidththree}{\treewidthfour}%
\addtolength{\branchwidththree}{\treeoffsetthree}%
\addtolength{\branchwidththree}{-\treeshiftthree}%
\addtolength{\branchwidththree}{-\treeoffsetfour}%
\setlength{\branchwidth}{\branchwidthone}%
\ifdim\branchwidthtwo>\branchwidth%
\setlength{\branchwidth}{\branchwidthtwo}\fi%
\ifdim\branchwidththree>\branchwidth%
\setlength{\branchwidth}{\branchwidththree}\fi%
\setlength{\daughteroffsetone}{\branchwidth}%
\addtolength{\daughteroffsetone}{-\branchwidthone}%
\addtolength{\daughteroffsetone}{-\treeshiftone}%
\setlength{\daughteroffsettwo}{\branchwidth}%
\addtolength{\daughteroffsettwo}{-\branchwidthtwo}%
\addtolength{\daughteroffsettwo}{-\treeshifttwo}%
\setlength{\daughteroffsetthree}{\branchwidth}%
\addtolength{\daughteroffsetthree}{-\branchwidththree}%
\addtolength{\daughteroffsetthree}{-\treeshiftthree}%
\setlength{\parentoffset}{-0.5\wd\parentbox}%
\addtolength{\parentoffset}{\treeoffsetfour}%
\addtolength{\parentoffset}{1.5\branchwidth}%
\setlength{\daughteroffset}{0in}%
\ifdim\parentoffset<0in%
\setlength{\daughteroffset}{-\parentoffset}%
\setlength{\parentoffset}{0in}\fi%
\setlength{\parentwidth}{\parentoffset}%
\addtolength{\parentwidth}{\wd\parentbox}%
\setlength{\treeoffset}{\daughteroffset}%
\addtolength{\treeoffset}{\treeoffsetfour}%
\setlength{\treewidth}{\wd\treeboxone}%
\addtolength{\treewidth}{\daughteroffsetone}%
\addtolength{\treewidth}{\treewidthtwo}%
\addtolength{\treewidth}{\daughteroffsettwo}%
\addtolength{\treewidth}{\treewidththree}%
\addtolength{\treewidth}{\daughteroffsetthree}%
\addtolength{\treewidth}{\treewidthfour}%
\addtolength{\treewidth}{\daughteroffset}%
\ifdim\treewidth<\parentwidth\setlength{\treewidth}{\parentwidth}\fi%
\sbox{\treebox}{\begin{minipage}{\treewidth}%
\begin{flushleft}%
\hspace*{\parentoffset}\usebox{\parentbox}\\
{\setlength{\unitlength}{0.5\branchwidth}%
\hspace*{\treeoffset}\begin{picture}(6,1)%
\put(0,0){\line(3,1){3}}%
\put(2,0){\line(1,1){1}}%
\put(4,0){\line(-1,1){1}}%
\put(6,0){\line(-3,1){3}}%
\end{picture}}\\
\vspace{-\baselineskip}
\hspace*{\daughteroffset}%
\makebox[\treewidthfour][l]%
{\raisebox{-\ht\treeboxfour}{\usebox{\treeboxfour}}}%
\hspace*{\daughteroffsetthree}%
\makebox[\treewidththree][l]%
{\raisebox{-\ht\treeboxthree}{\usebox{\treeboxthree}}}%
\hspace*{\daughteroffsettwo}%
\makebox[\treewidthtwo][l]%
{\raisebox{-\ht\treeboxtwo}{\usebox{\treeboxtwo}}}%
\hspace*{\daughteroffsetone}%
\raisebox{-\ht\treeboxone}{\usebox{\treeboxone}}%
\end{flushleft}%
\end{minipage}}%
\setlength{\treeoffsetone}{\parentoffset}%
\addtolength{\treeoffsetone}{0.5\wd\parentbox}%
\setlength{\treeshiftone}{0pt}%
\setlength{\treewidthone}{\treewidth}%
\sbox{\treeboxone}{\usebox{\treebox}}\poptree\poptree\poptree%
\else\ifnum\value{branchcount}=5\sbox{\parentbox}{\ontop{#2}}%
\setlength{\branchwidthone}{\treewidthtwo}%
\addtolength{\branchwidthone}{\treeoffsetone}%
\addtolength{\branchwidthone}{-\treeshiftone}%
\addtolength{\branchwidthone}{-\treeoffsettwo}%
\setlength{\branchwidthtwo}{\treewidththree}%
\addtolength{\branchwidthtwo}{\treeoffsettwo}%
\addtolength{\branchwidthtwo}{-\treeshifttwo}%
\addtolength{\branchwidthtwo}{-\treeoffsetthree}%
\setlength{\branchwidththree}{\treewidthfour}%
\addtolength{\branchwidththree}{\treeoffsetthree}%
\addtolength{\branchwidththree}{-\treeshiftthree}%
\addtolength{\branchwidththree}{-\treeoffsetfour}%
\setlength{\branchwidthfour}{\treewidthfive}%
\addtolength{\branchwidthfour}{\treeoffsetfour}%
\addtolength{\branchwidthfour}{-\treeshiftfour}%
\addtolength{\branchwidthfour}{-\treeoffsetfive}%
\setlength{\branchwidth}{\branchwidthone}%
\ifdim\branchwidthtwo>\branchwidth%
\setlength{\branchwidth}{\branchwidthtwo}\fi%
\ifdim\branchwidththree>\branchwidth%
\setlength{\branchwidth}{\branchwidththree}\fi%
\ifdim\branchwidthfour>\branchwidth%
\setlength{\branchwidth}{\branchwidthfour}\fi%
\setlength{\daughteroffsetone}{\branchwidth}%
\addtolength{\daughteroffsetone}{-\branchwidthone}%
\addtolength{\daughteroffsetone}{-\treeshiftone}%
\setlength{\daughteroffsettwo}{\branchwidth}%
\addtolength{\daughteroffsettwo}{-\branchwidthtwo}%
\addtolength{\daughteroffsettwo}{-\treeshifttwo}%
\setlength{\daughteroffsetthree}{\branchwidth}%
\addtolength{\daughteroffsetthree}{-\branchwidththree}%
\addtolength{\daughteroffsetthree}{-\treeshiftthree}%
\setlength{\daughteroffsetfour}{\branchwidth}%
\addtolength{\daughteroffsetfour}{-\branchwidthfour}%
\addtolength{\daughteroffsetfour}{-\treeshiftfour}%
\setlength{\parentoffset}{-0.5\wd\parentbox}%
\addtolength{\parentoffset}{\treeoffsetfive}%
\addtolength{\parentoffset}{2\branchwidth}%
\setlength{\daughteroffset}{0in}%
\ifdim\parentoffset<0in%
\setlength{\daughteroffset}{-\parentoffset}%
\setlength{\parentoffset}{0in}\fi%
\setlength{\parentwidth}{\parentoffset}%
\addtolength{\parentwidth}{\wd\parentbox}%
\setlength{\treeoffset}{\daughteroffset}%
\addtolength{\treeoffset}{\treeoffsetfive}%
\setlength{\treewidth}{\wd\treeboxone}%
\addtolength{\treewidth}{\daughteroffsetone}%
\addtolength{\treewidth}{\treewidthtwo}%
\addtolength{\treewidth}{\daughteroffsettwo}%
\addtolength{\treewidth}{\treewidththree}%
\addtolength{\treewidth}{\daughteroffsetthree}%
\addtolength{\treewidth}{\treewidthfour}%
\addtolength{\treewidth}{\daughteroffsetfour}%
\addtolength{\treewidth}{\treewidthfive}%
\addtolength{\treewidth}{\daughteroffset}%
\ifdim\treewidth<\parentwidth\setlength{\treewidth}{\parentwidth}\fi%
\sbox{\treebox}{\begin{minipage}{\treewidth}%
\begin{flushleft}%
\hspace*{\parentoffset}\usebox{\parentbox}\\
{\setlength{\unitlength}{0.5\branchwidth}%
\hspace*{\treeoffset}\begin{picture}(8,1)%
\put(0,0){\line(4,1){4}}%
\put(2,0){\line(2,1){2}}%
\put(4,0){\line(0,1){1}}%
\put(6,0){\line(-2,1){2}}%
\put(8,0){\line(-4,1){4}}%
\end{picture}}\\
\vspace{-\baselineskip}
\hspace*{\daughteroffset}%
\makebox[\treewidthfive][l]%
{\raisebox{-\ht\treeboxfour}{\usebox{\treeboxfive}}}%
\hspace*{\daughteroffsetfour}%
\makebox[\treewidthfour][l]%
{\raisebox{-\ht\treeboxfour}{\usebox{\treeboxfour}}}%
\hspace*{\daughteroffsetthree}%
\makebox[\treewidththree][l]%
{\raisebox{-\ht\treeboxthree}{\usebox{\treeboxthree}}}%
\hspace*{\daughteroffsettwo}%
\makebox[\treewidthtwo][l]%
{\raisebox{-\ht\treeboxtwo}{\usebox{\treeboxtwo}}}%
\hspace*{\daughteroffsetone}%
\raisebox{-\ht\treeboxone}{\usebox{\treeboxone}}%
\end{flushleft}%
\end{minipage}}%
\setlength{\treeoffsetone}{\parentoffset}%
\addtolength{\treeoffsetone}{0.5\wd\parentbox}%
\setlength{\treeshiftone}{0pt}%
\setlength{\treewidthone}{\treewidth}%
\sbox{\treeboxone}{\usebox{\treebox}}\poptree\poptree\poptree\poptree%
\else\typeout{QobiTeX warning--- Can't handle #1 branching}\fi\fi\fi\fi\fi}
\newcommand{\faketreewidth}[1]{%
\sbox{\parentbox}{\ontop{#1}}%
\setlength{\treewidthone}{0.5\wd\parentbox}%
\addtolength{\treewidthone}{\treeoffsetone}%
\setlength{\treeshiftone}{\treeoffsetone}%
\addtolength{\treeshiftone}{-0.5\wd\parentbox}}
\newcommand{\tree}{%
\usebox{\treeboxone}
\setlength{\treeoffsetone}{\treeoffsettwo}%
\sbox{\treeboxone}{\usebox{\treeboxtwo}}%
\poptree}

\input{psfig}
\documentstyle[aclap]{article}

\newcommand{\derives}{\stackrel{\ast}{\Rightarrow}}

\title{{\normalsize \tt To appear in {\it Proceedings of the 34th Annual
Meeting of the ACL}, June 1996} \\ \mbox{} \\
Parsing Algorithms and Metrics}

\author{Joshua Goodman\\
        Harvard University\\
	33 Oxford St. \\
        Cambridge, MA 02138\\
        goodman@das.harvard.edu}

\begin{document}
\maketitle
\bibliographystyle{fullname}

\begin{abstract}
Many different metrics exist for evaluating parsing results, including
Viterbi, Crossing Brackets Rate, Zero Crossing Brackets Rate, and
several others.  However, most parsing algorithms, including the
Viterbi algorithm, attempt to optimize the same metric, namely the
probability of getting the correct labelled tree.  By choosing a
parsing algorithm appropriate for the evaluation metric, better
performance can be achieved.  We present two new algorithms: the
``Labelled Recall Algorithm,'' which maximizes the expected Labelled
Recall Rate, and the ``Bracketed Recall Algorithm,'' which maximizes
the Bracketed Recall Rate.  Experimental results are given, showing
that the two new algorithms have improved performance over the Viterbi
algorithm on many criteria, especially the ones that they optimize.  
\end{abstract}

\section{Introduction}

In corpus-based approaches to parsing, one is given a treebank (a
collection of text annotated with the ``correct'' parse tree) and
attempts to find algorithms that, given unlabelled text from the
treebank, produce as similar a parse as possible to the one in the
treebank.

Various methods can be used for finding these parses.  Some of the
most common involve inducing Probabilistic Context-Free Grammars
(PCFGs), and then parsing with an algorithm such as the Labelled Tree
(Viterbi) Algorithm, which maximizes the probability that the output
of the parser (the ``guessed'' tree) is the one that the PCFG
produced.  This implicitly assumes that the induced PCFG does a good
job modeling the corpus.

There are many different ways to evaluate these parses.  The most
common include the Labelled Tree Rate (also called the Viterbi
Criterion or Exact Match Rate), Consistent Brackets Recall Rate
(also called the Crossing Brackets Rate), Consistent Brackets Tree
Rate (also called the Zero Crossing Brackets Rate), and Precision
and Recall.  Despite the variety of evaluation metrics, nearly all
researchers use algorithms that maximize performance on the Labelled
Tree Rate, even in domains where they are evaluating using other
criteria.

We propose that by creating algorithms that optimize the evaluation
criterion, rather than some related criterion, improved performance
can be achieved.  

In Section \ref{sec:evalmet}, we define most of the evaluation
metrics used in this paper and discuss previous approaches.  Then, in
Section \ref{sec:labconspars}, we discuss the Labelled Recall
Algorithm, a new algorithm that maximizes performance on the Labelled
Recall Rate.  In Section \ref{sec:brackconspars}, we discuss another
new algorithm, the Bracketed Recall Algorithm, that maximizes
performance on the Bracketed Recall Rate (closely related to the
Consistent Brackets Recall Rate).  Finally, we give experimental
results in Section \ref{sec:results1} using these two algorithms in
appropriate domains, and compare them to the Labelled Tree (Viterbi)
Algorithm, showing that each algorithm generally works best when
evaluated on the criterion that it optimizes.

\section{Evaluation Metrics}
\label{sec:evalmet}

In this section, we first define basic terms and symbols.  Next, we
define the different metrics used in evaluation.  Finally, we discuss
the relationship of these metrics to parsing algorithms.

\subsection{Basic Definitions}

Let $w_a$ denote word $a$ of the sentence under consideration.  Let
$w_a^b$ denote $w_a w_{a+1} ... w_{b-1} w_b$; in particular let
$w_1^n$ denote the entire sequence of terminals (words) in the
sentence under consideration.

In this paper we assume all guessed parse trees are binary branching.  Let a
parse tree T be defined as a set of triples $(s, t, X)$---where $s$
denotes the position of the first symbol in a constituent, $t$ denotes
the position of the last symbol, and $X$ represents a terminal or
nonterminal symbol---meeting the following three requirements:  
\begin{itemize}

\item The sentence was generated by the start
symbol, $S$.  Formally, $(1, n, S) \in T$.
\item Every word in the sentence is in the parse tree.  Formally,
for every $s$ between 1 and $n$ the triple $(s, s, w_s) \in T$.
\item The tree is binary branching and consistent.  Formally, for
every $(s, t, X)$ in $T$, $s \neq t$, there is exactly one $r, Y,$ and $Z$
such that $s \leq r < t$ and $(s, r, Y) \in T$ and $(r+1, t, Z) \in T$.
\end{itemize}

Let $T_C$ denote the ``correct'' parse (the one in the treebank) and
let $T_G$ denote the ``guessed'' parse (the one output by the parsing
algorithm).  Let $N_G$ denote $\vert T_G \vert$, the number of
nonterminals in the guessed parse tree, and let $N_C$ denote $\vert
T_C \vert$, the number of nonterminals in the correct parse tree.

\subsection{Evaluation Metrics}
There are various levels of strictness for determining whether a
constituent (element of $T_G$) is ``correct.''  The strictest of these
is Labelled Match.  A constituent $(s, t, X) \in T_G$ is correct
according to Labelled Match if and only if $(s, t, X) \in T_C.$ In
other words, a constituent in the guessed parse tree is correct if and only if it
occurs in the correct parse tree.

The next level of strictness is Bracketed Match.  Bracketed match is
like labelled match, except that the nonterminal label is ignored.
Formally, a constituent $(s, t, X) \in T_G$ is correct according to
Bracketed Match if and only if there exists a $Y$ such that $(s, t, Y)
\in T_C.$

The least strict level is Consistent Brackets (also called Crossing
Brackets).  Consistent Brackets is like Bracketed Match in that the
label is ignored.  It is even less strict in that the observed $(s, t,
X)$ need not be in $T_C$---it must simply not be ruled out by any $(q,
r, Y) \in T_C$.  A particular triple $(q, r, Y)$ rules out $(s, t, X)$
if there is no way that $(s, t, X)$ and $(q, r, Y)$ could both be in
the same parse tree.  In particular, if the interval $(s, t)$ crosses
the interval $(q, r)$, then $(s, t, X)$ is ruled out and counted as an
error.  Formally, we say that $(s, t)$ crosses $(q, r)$ if and only if
$s < q \leq t < r$ or $q < s \leq r <  t$.
%\footnote{This follows Pereira and Schabes
%\cite{Pereira:92a}, except that since in our notation spans include the first
%element, the inequalities are slightly different.}

If $T_C$ is binary branching, then Consistent Brackets and Bracketed
Match are identical.  The following symbols denote the number of
constituents that match according to each of these criteria.

\begin{list}{}
\item $L = \vert T_C \cap T_G \vert$ : the number of constituents in $T_G$
that are correct according to Labelled Match.

\item $B = \vert \{(s, t, X) : (s, t, X) \in T_G$ and for some
$Y\: (s, t, Y) \in T_C\}\vert $: the number of constituents in $T_G$
that are correct according to Bracketed Match.

\item $C = \vert \{(s, t, X) \in T_G : $ there is no $(v, w, Y)
\in T_C$ crossing $(s, t)\} \vert$ : the number of constituents
in $T_G$ correct according to Consistent Brackets.

\end{list}

\smallskip

Following are the definitions of the six metrics used in this paper for
evaluating binary branching trees:

\newcounter{list:metrics}

\newcommand{\oneif}{\mbox{ 1 if }}

\begin{list} {(\arabic{list:metrics})}{\usecounter{list:metrics}}

\item {\em Labelled Recall Rate} = $L/N_C$.

\item {\em Labelled Tree Rate} = $\oneif L = N_C$.  It is also called 
the Viterbi Criterion.

\item {\em Bracketed Recall Rate} = $B/N_C$.

\item {\em Bracketed Tree Rate} = $\oneif B = N_C$.

\item {\em Consistent Brackets Recall Rate} = $C/N_G$.  It is often
called the Crossing Brackets Rate.  In the case where the parses are
binary branching, this criterion is the same as the Bracketed Recall Rate.

\item {\em Consistent Brackets Tree Rate} = $\oneif C = N_G$.  This
metric is closely related to the Bracketed Tree Rate.  In the case
where the parses are binary branching, the two metrics are the same.
This criterion is also called the Zero Crossing Brackets Rate.

\end{list}

\noindent The preceding six metrics each correspond to cells in the following
table:

\smallskip

\begin{center}
\begin{tabular}{|l||r|r|} \hline
	          & Recall         & Tree                \\ \hline
                                                              \hline
Consistent Brackets & $C / N_G$            & 1 if $C = N_G$     \\
\hline
Brackets          & $B / N_C$            & 1 if $B = N_C$     \\
\hline
Labels            &  $L / N_C$           & 1 if $L = N_C$      \\
\hline
\end{tabular}
\end{center}

\mbox{}

\subsection {Maximizing Metrics}

Despite this long list of possible metrics, there is only one metric
most parsing algorithms attempt to maximize, namely the Labelled Tree
Rate.  That is, most parsing algorithms assume that the test corpus
was generated by the model, and then attempt to evaluate the following
expression, where $E$ denotes the expected value operator:
\begin{equation}
\label{eqn:labtremax}
T_G=\arg \max_T E(\oneif L = N_C)
\end{equation}
This is true of the Labelled Tree Algorithm and stochastic versions of
Earley's Algorithm \cite{Stolcke:93a}, and variations such as those
used in Picky parsing \cite{Magerman:92a}.  Even in probabilistic
models not closely related to PCFGs, such as Spatter parsing
\cite{Magerman:94a}, expression (\ref{eqn:labtremax}) is still
computed.  One notable exception is Brill's Transformation-Based Error
Driven system \cite{Brill:93a}, which induces a set of transformations
designed to maximize the Consistent Brackets Recall Rate.  However, Brill's
system is not probabilistic.  Intuitively, if one were to match the
parsing algorithm to the evaluation criterion, better performance should
be achieved.

Ideally, one might try to directly maximize the most commonly used
evaluation criteria, such as Consistent Brackets Recall (Crossing
Brackets) Rate.  Unfortunately, this criterion is relatively difficult to
maximize, since it is time-consuming to compute the probability that a
particular constituent crosses some constituent in the correct parse.
On the other hand, the Bracketed Recall and Bracketed Tree Rates are
easier to handle, since computing the probability that a bracket
matches one in the correct parse is inexpensive.  It is plausible that
algorithms which optimize these closely related criteria will do well
on the analogous Consistent Brackets criteria.

\subsection{Which Metrics to Use} 
When building an actual system, one should use the metric most
appropriate for the problem.  For instance, if one were creating a
database query system, such as an ATIS system, then the Labelled Tree
(Viterbi) metric would be most appropriate.  A single error in the
syntactic representation of a query will likely result in an error in
the semantic representation, and therefore in an incorrect database
query, leading to an incorrect result.  For instance, if the user
request ``Find me all flights on Tuesday'' is misparsed with the
prepositional phrase attached to the verb, then the system might wait
until Tuesday before responding: a single error leads to completely
incorrect behavior.  Thus, the Labelled Tree criterion is appropriate.

On the other hand, consider a machine assisted translation system, in
which the system provides translations, and then a fluent human
manually edits them.  Imagine that the system is given the foreign
language equivalent of ``His credentials are nothing which should be
laughed at,'' and makes the single mistake of attaching the relative
clause at the sentential level, translating the sentence as ``His
credentials are nothing, which should make you laugh.''  While the
human translator must make some changes, he certainly needs to do less
editing than he would if the sentence were completely misparsed.  The
more errors there are, the more editing the human translator needs to
do.  Thus, a criterion such as the Labelled Recall criterion is
appropriate for this task, where the number of incorrect constituents
correlates to application performance.

\section{Labelled Recall Parsing}
\label{sec:labconspars}

Consider writing a parser for a domain such as machine assisted
translation.  One could use the Labelled Tree Algorithm, which would
maximize the expected number of exactly correct parses.  However,
since the number of correct constituents is a better measure of
application performance for this domain than the number of correct
trees, perhaps one should use an algorithm which maximizes the
Labelled Recall criterion, rather than the Labelled Tree criterion.

The Labelled Recall Algorithm finds that tree $T_G$ which has the
highest expected value for the Labelled Recall Rate, $L/N_C$ (where
$L$ is the number of correct labelled constituents, and $N_C$ is the
number of nodes in the correct parse).  This can be written as
follows:

\begin{equation} \label{eqn:max1}
T_G=\arg \max_T E(L/N_C)
\end{equation}

It is not immediately obvious that the maximization of expression
(\ref{eqn:max1}) is in fact different from the maximization of expression
(\ref{eqn:labtremax}), but a simple example illustrates the
difference.  The following grammar generates four trees with equal
probability:
\begin{equation}
\begin{array}{rcll}
\mbox{S} & \rightarrow  & \mbox{A C}  & 0.25 \\
\mbox{S} & \rightarrow  & \mbox{A D}  & 0.25 \\
\mbox{S} & \rightarrow  & \mbox{E B}  & 0.25 \\
\mbox{S} & \rightarrow  & \mbox{F B}  & 0.25 \\
\mbox{A, B, C, D, E, F} & \rightarrow & \mbox {x x} & 1.0  \\
\end{array}
\end{equation}
The four trees are
%
% Use faketreewidth to work around a bug
%
\begin{center}
\begin{tabular}{cc}
\leaf{x} \faketreewidth{AA}
\leaf{x} \faketreewidth{AA}
\branch{2}{A}
\leaf{x} \faketreewidth{AA}
\leaf{x} \faketreewidth{AA}
\branch{2}{C}
\faketreewidth{AA}
\branch{2}{S}
\faketreewidth{SS}
\hspace{-0.5in} \tree \hspace{-0.1in} \vspace{0.5em}&
\leaf{x} \faketreewidth{AA}
\leaf{x} \faketreewidth{AA}
\branch{2}{A}
\leaf{x} \faketreewidth{AA}
\leaf{x} \faketreewidth{AA}
\branch{2}{D}
\faketreewidth{AA}
\branch{2}{S}
\faketreewidth{SS}
\hspace{-0.5in} \tree \hspace{-0.1in} \vspace{0.5em} \\
\leaf{x} \faketreewidth{AA}
\leaf{x} \faketreewidth{AA}
\branch{2}{E}
\leaf{x} \faketreewidth{AA}
\leaf{x} \faketreewidth{AA}
\branch{2}{B}
\faketreewidth{AA}
\branch{2}{S}
\faketreewidth{SS}
\hspace{-0.5in} \tree \hspace{-0.1in} &
\leaf{x} \faketreewidth{AA}
\leaf{x} \faketreewidth{AA}
\branch{2}{F}
\leaf{x} \faketreewidth{AA}
\leaf{x} \faketreewidth{AA}
\branch{2}{B}
\faketreewidth{AA}
\branch{2}{S}
\faketreewidth{SS}
\hspace{-0.5in} \tree \hspace{-0.1in}
\end{tabular}
\end{center}

For the first tree, the probabilities of being correct are S: 100\%;
A:50\%; and C: 25\%.  Similar counting holds for the other three.
Thus, the expected value of $L$ for any of these trees is 1.75.

On the other hand, the optimal Labelled Recall parse is
\leaf{x} \faketreewidth{AA}
\leaf{x} \faketreewidth{AA}
\branch{2}{A}
\leaf{x} \faketreewidth{AA}
\leaf{x} \faketreewidth{AA}
\branch{2}{B}
\faketreewidth{SS}
\branch{2}{S}
\begin{center}
\tree
\end{center}
This tree has 0 probability according to the grammar, and thus is
non-optimal according to the Labelled Tree Rate criterion.  However,
for this tree the probabilities of each node being correct are S:
100\%; A: 50\%; and B: 50\%.  The expected value of $L$ is 2.0, the
highest of any tree.  This tree therefore optimizes the Labelled
Recall Rate.

\subsection{Algorithm}
We now derive an algorithm for finding the parse that maximizes the
expected Labelled Recall Rate.  We do this by expanding
expression (\ref{eqn:max1}) out into a probabilistic
form, converting this into a recursive equation, and finally creating
an equivalent dynamic programming algorithm.

We begin by rewriting expression (\ref{eqn:max1}), expanding out the
expected value operator, and removing the $\frac{1}{N_C}$, which is
the same for all $T_G$, and so plays no role in the maximization.

\begin{equation} \label{eqn:max2}
T_G=\arg \max_T \sum_{T_C} P(T_C \mid w_1^n) \; |T \cap T_C|
\end{equation}

This can be further expanded to

\begin{equation} \label{eqn:max3}
T_G=\arg \max_T \! \sum_{T_C} P(T_C\mid w_1^n) \hspace{-1.5em}
\sum_{(s, t, X) \in T} \hspace{-1.5em} \oneif (s, t, X) \in T_C
\end{equation}

Now, given a PCFG with start symbol $S$, the following equality
holds:

\begin{eqnarray}
\lefteqn{P(S \derives w_1^{s-1} X w_{t+1}^n \mid w_1^n) =} & & \nonumber\\
 & &\sum_{T_C} P(T_C\mid w_1^n) (\oneif (s, t, X) \in T_C)
\end{eqnarray}

By rearranging the summation in expression (\ref{eqn:max3}) and then  
substituting this equality, we get

\begin{equation}
T_G=\arg \max_T \sum_{(s, t, X) \in T}
P(S \derives w_1^{s-1} X w_{t+1}^n \mid w_1^n)
\end{equation}

At this point, it is useful to introduce the Inside and Outside
probabilities, due to \newcite{Baker:79b}, and explained by
\newcite{Lari:90a}.  The Inside probability is defined as $e(s, t, X)
= P(X \derives w_s^t)$ and the Outside probability is $f(s, t, X) =
P(S \derives w_1^{s-1} X w_{t+1}^n)$.  Note that while Baker and
others have used these probabilites for inducing grammars, here they
are used only for parsing.

Let us define a new function, $g(s, t, X)$.

\begin{eqnarray*}
g(s, t, X)  & = &  P(S \derives w_1^{s-1} X w_{t+1}^n \mid w_1^n)\\
   &=& \frac{P(S \derives w_1^{s-1} X w_{t+1}^n ) 
                     P(X \derives w_s^{t}) }
                    { P(S \derives w_1^n) } \\
   &=& f(s, t, X) \times e(s, t, X) / e(1, n, S)
\end{eqnarray*}

Now, the definition of a Labelled Recall Parse can be rewritten
as

\begin{equation}
T_G=\arg \max_T \sum_{(s, t, X) \in T} g(s, t, X)
\end{equation}

\def\MAXC{\mathop{\rm MAXC}\nolimits}

Given the matrix $g(s, t, X)$, it is a simple matter of dynamic
programming to determine the parse that maximizes the Labelled Recall
criterion.  Define

\begin{eqnarray}
\lefteqn{\MAXC(s, t)  =\max_{X} g(s, t, X) + } & & \nonumber \\
	     &   &\max_{r \mid s \leq r < t}\left( \MAXC(s, r)+\MAXC(r+1, t) \right) \nonumber
\label{eqn:maxc}
\end{eqnarray}

It is clear that $\MAXC(1, n)$ contains the score of the best parse
according to the Labelled Recall criterion.  This equation can
be converted into the dynamic programming algorithm shown in Figure
\ref{fig:maxcons}.
\begin{figure}
\begin{verbatim}
for length := 2 to n
   for s := 1 to n-length+1
      t := s + length - 1;
      loop over nonterminals X
         let max_g:=maximum of g(s,t,X)
      loop over r such that s <= r < t
         let best_split:=
            max of maxc[s,r] + maxc[r+1,t]
      maxc[s, t] := max_g + best_split;
\end{verbatim}
\caption{Labelled Recall Algorithm} \label{fig:maxcons}
\end{figure}

For a grammar with $r$ rules and $k$ nonterminals, the run time of
this algorithm is $O(n^3 + kn^2)$ since there are two layers
of outer loops, each with run time at most $n$, and an inner loop,
over nonterminals and $n$.  However, this is dominated by the
computation of the Inside and Outside probabilities, which takes time
$O(rn^3)$.

By modifying the algorithm slightly to record the actual split used at each
node, we can recover the best parse.  The entry {\tt maxc[1, n]}
contains the expected number of correct constituents, given the model.

\section{Bracketed Recall Parsing}
\label{sec:brackconspars}

The Labelled Recall Algorithm maximizes the expected number of correct
labelled constituents.  However, many commonly used evaluation
metrics, such as the Consistent Brackets Recall Rate, ignore labels.
Similarly, some grammar induction algorithms, such as those used by
\newcite{Pereira:92a} do not produce meaningful
labels.  In particular, the Pereira and Schabes method induces a
grammar from the brackets in the treebank, ignoring the labels.  While
the induced grammar has labels, they are not related to those in the
treebank.  Thus, although the Labelled Recall Algorithm could be used in
these domains, perhaps maximizing a criterion that is more closely
tied to the domain will produce better results.  Ideally, we would
maximize the Consistent Brackets Recall Rate directly.  However, since it is
time-consuming to deal with Consistent Brackets, we instead use the
closely related Bracketed Recall Rate.

For the Bracketed Recall Algorithm, we find the parse that
maximizes the expected Bracketed Recall Rate, $B/N_C$.  (Remember that
$B$ is the number of brackets that are correct, and $N_C$ is the
number of constituents in the correct parse.)

\begin{equation}
T_G=\arg \max_T E(B/N_C)
\label{eqn:maxbracket}
\end{equation}

Following a derivation similar to that used for the Labelled Recall
Algorithm, we can rewrite equation (\ref{eqn:maxbracket}) as

\begin{equation}
T_G=\arg \max_T \sum_{(s, t) \in T} \sum_{X}
P(S \derives w_1^{s-1} X w_{t+1}^n \mid w_1^n)
\end{equation}

The algorithm for Bracketed Recall parsing is extremely similar to
that for Labelled Recall parsing.  The only required change is that we
sum over the symbols $X$ to calculate {\tt  max\_g}, rather than maximize
over them.

\section{Experimental Results}
\label{sec:results1}

We describe two experiments for testing these algorithms.  The first
uses a grammar without meaningful nonterminal symbols, and compares
the Bracketed Recall Algorithm to the traditional Labelled Tree
(Viterbi) Algorithm.  The second uses a grammar with meaningful
nonterminal symbols and performs a three-way comparison between the
Labelled Recall, Bracketed Recall, and Labelled Tree Algorithms.
These experiments show that use of an algorithm matched appropriately
to the evaluation criterion can lead to as much as a 10\% reduction in
error rate.

In both experiments the grammars could not parse some sentences, 0.5\% and
9\%, respectively.  The unparsable data were assigned a right
branching structure with their rightmost element attached high.  Since
all three algorithms fail on the same sentences, all algorithms were
affected equally.

\subsection{Experiment with Grammar Induced by Pereira and Schabes Method}
The experiment of \newcite{Pereira:92a} was
duplicated.  In that experiment, a grammar was trained from a
bracketed form of the TI section of the ATIS corpus\footnote{For our
experiments the corpus was slightly cleaned up.  A diff file for
``ed'' between the original ATIS data and the cleaned-up version is
available from 
ftp://ftp.das.harvard.edu/pub/goodman/atis-ed/
ti\_tb.par-ed and ti\_tb.pos-ed.
The number of changes made was
small, less than 0.2\%}
using a modified form of the Inside-Outside Algorithm.  Pereira and
Schabes then used the Labelled Tree Algorithm to select the best parse
for sentences in held out test data.  The experiment was repeated
here, except that both the Labelled Tree and Labelled Recall Algorithm
were run for each sentence.  In contrast to previous research, we
repeated the experiment ten times, with different training set, test
set, and initial conditions each time.

\begin{table}
\begin{center}
\begin {tabular}{|l||r|r|r|r|} \hline
           Criteria & Min    & Max    & Mean  & SDev  \\ \hline
\hline \multicolumn{5}{|c|} {Labelled Tree Algorithm} \\ \hline
Cons Brack Rec &86.06 &93.27 &90.13 & 2.57  \\ \hline
Cons Brack Tree   &51.14 &77.27 &63.98 & 7.96  \\ \hline
Brack Rec      &71.38 &81.88 &75.87 & 3.18  \\ \hline
\hline \multicolumn{5}{|c|} {Bracketed Recall Algorithm} \\ \hline
Cons Brack Rec &88.02 &94.34 &91.14 & 2.22  \\ \hline
Cons Brack Tree   &53.41 &76.14 &63.64 & 7.82  \\ \hline
Brack Rec      &72.15 &80.69 &76.03 & 3.14  \\ \hline
\hline \multicolumn{5}{|c|} {Differences} \\ \hline	  
Cons Brack Rec &-1.55 & 2.45 & 1.01 & 1.07  \\ \hline
Cons Brack Tree   &-3.41 & 3.41 &-0.34 & 2.34  \\ \hline
Brack Rec      &-1.34 & 2.02 & 0.17 & 1.20  \\ \hline
\end{tabular}
\end{center}
\caption{Percentages Correct for Labelled Tree versus Bracketed Recall
for Pereira and Schabes}
\label{tab:maxbrackcompare}
\end{table}

Table \ref{tab:maxbrackcompare} shows the results of running this
experiment, giving the minimum, maximum, mean, and standard deviation
for three criteria, Consistent Brackets Recall, Consistent Brackets
Tree, and Bracketed Recall.  We also display these statistics for the
paired differences between the algorithms.  The only statistically
significant difference is that for Consistent Brackets Recall Rate,
which was significant to the 2\% significance level (paired t-test).
Thus, use of the Bracketed Recall Algorithm leads to a 10\% reduction
in error rate.

\begin{figure*}
\psfig{figure=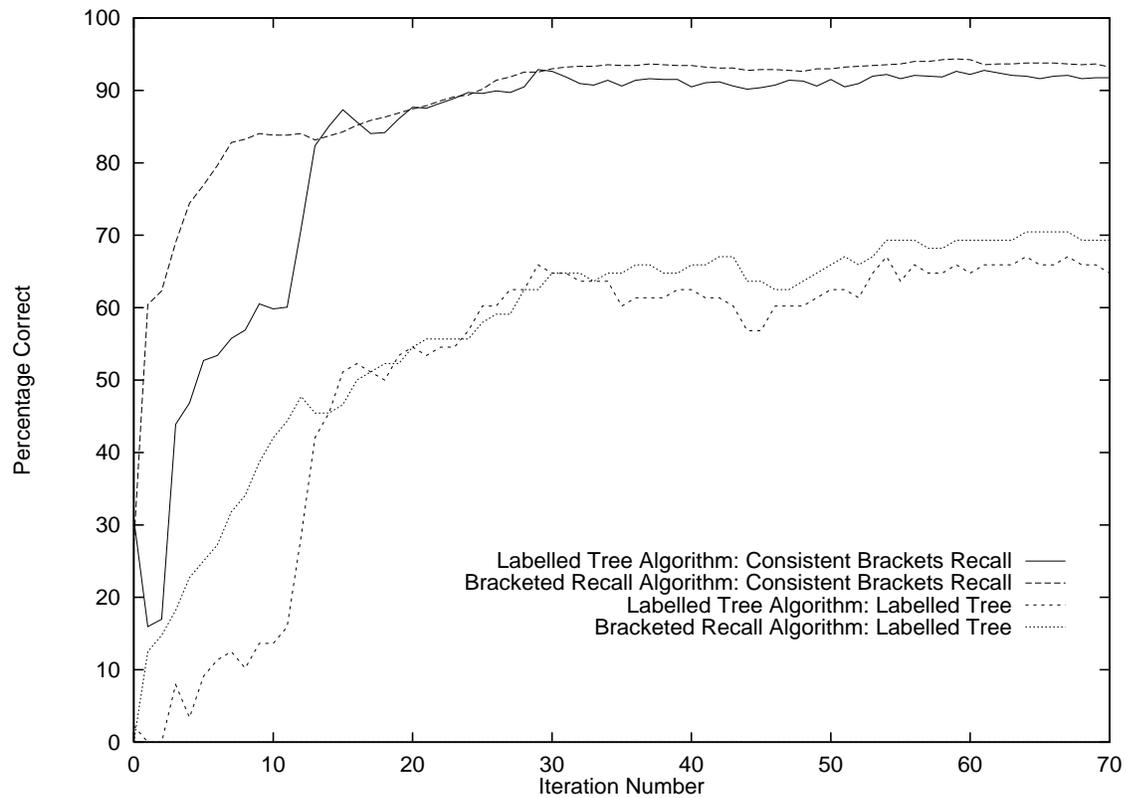,width=6in}

\caption{Labelled Tree versus Bracketed Recall in Pereira and Schabes
Grammar} \label{fig:typpns}
\end{figure*}

In addition, the performance of the Bracketed Recall Algorithm was
also qualitatively more appealing.  Figure \ref{fig:typpns} shows
typical results.  Notice that the Bracketed Recall Algorithm's
Consistent Brackets Rate (versus iteration) is smoother and more
nearly monotonic than the Labelled Tree Algorithm's. The Bracketed
Recall Algorithm also gets off to a much faster start, and is generally
(although not always) above the Labelled Tree level.  For the Labelled
Tree Rate, the two are usually very comparable.

\subsection{Experiment with Grammar Induced by Counting}
The replication of the Pereira and Schabes experiment was useful for
testing the Bracketed Recall Algorithm.  However, since that
experiment induces a grammar with nonterminals not comparable to
those in the training, a different experiment is needed to evaluate
the Labelled Recall Algorithm, one in which the nonterminals in
the induced grammar are the same as the nonterminals in the test set.

\subsubsection{Grammar Induction by Counting}
\begin{figure}
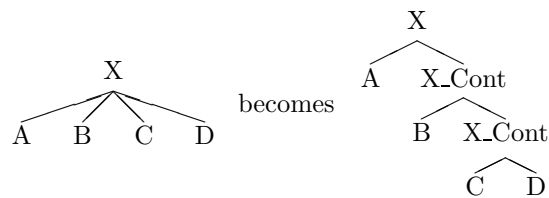

\hspace{-0.3in}
\mbox{
\leaf{A}
\leaf{B} \faketreewidth{SS}
\leaf{C} \faketreewidth{SS}
\leaf{D}
\branch{4}{X}
\tree
\hspace{-0.1in}
} becomes
\mbox{
\leaf{A}
\leaf{B}
\leaf{C} \faketreewidth{SS}
\leaf{D} \faketreewidth{SS}
\branch{2}{X\_Cont}
\branch{2}{X\_Cont}
\branch{2}{X}
\hspace{-0.5in}
\tree
\hspace{-0.3in}
}
\caption{Conversion of Productions to Binary Branching}
\label{fig:binary}
\end{figure}

For this experiment, a very simple grammar was induced by counting,
using a portion of the Penn Tree Bank, version 0.5.
% Should I cite someone here?  Who?
In particular, the trees were first made binary branching by removing
epsilon productions, collapsing singleton productions, and converting
n-ary productions ($n > 2$) as in figure \ref{fig:binary}.  The
resulting trees were treated as the ``Correct'' trees in the
evaluation.  Only trees with forty or fewer symbols were used in this
experiment.
% results in ~/cellar/wsj/parser-out73

A grammar was then induced in a straightforward way from these trees,
simply by giving one count for each observed production.  No smoothing
was done.  There were 1805 sentences and 38610 nonterminals in the
test data.  

\subsubsection{Results}
\begin{table*}
\begin{center}
\begin{tabular}{|l||r|r|r||r|r|} \cline{2-6}
\multicolumn{1}{c}{} &\multicolumn{5}{|c|} {Criterion} \\ \hline
               &    Label &    Label  &    Brack   &Cons Brack&Cons Brack\\ 
Algorithm      &    Tree  &    Recall &    Recall  &Recall    &Tree      \\ \hline\hline
Label Tree     &\em 4.54\%&    48.60\%&    60.98\% &66.35\%   &12.07\%   \\ \hline
Label Recall   &    3.71\%&\em 49.66\%&    61.34\% &68.39\%   &11.63\%   \\ \hline
Bracket Recall &    0.11\%&     4.51\%&\em 61.63\% &68.17\%   &11.19\%
\\ \hline
\end{tabular}
\end{center}
\caption{Grammar Induced by Counting: Three Algorithms 
Evaluated on Five Criteria} \label{tab:counting}
\end{table*}

Table \ref{tab:counting} shows the results of running all three
algorithms, evaluating against five criteria.  Notice that for
each algorithm, for the criterion that it optimizes it is the best
algorithm.  That is, the Labelled Tree Algorithm is the best for the
Labelled Tree Rate, the Labelled Recall Algorithm is
the best for the Labelled Recall Rate, and the Bracketed Recall
Algorithm is the best for the Bracketed Recall Rate.

\section{Conclusions and Future Work}

\smallskip

\begin{table}
\begin{center}
\begin{tabular}{|l||r|r|} \hline
	          & Recall            & Tree             \\ \hline \hline
Brackets          & Bracketed Recall  & (NP-Complete)    \\ \hline
Labels            & Labelled Recall   &  Labelled Tree   \\ \hline
\end{tabular}
\end{center}
\caption{Metrics and Corresponding Algorithms} \label{tab:ressum}
\end{table}

Matching parsing algorithms to evaluation criteria is a powerful
technique that can be used to improve performance.  In particular, the
Labelled Recall Algorithm can improve performance versus the Labelled
Tree Algorithm on the Consistent Brackets, Labelled Recall, and
Bracketed Recall criteria.  Similarly, the Bracketed Recall Algorithm
improves performance (versus Labelled Tree) on Consistent Brackets
and Bracketed Recall criteria.  Thus, these algorithms improve
performance not only on the measures that they were designed for, but
also on related criteria.

Furthermore, in some cases these techniques can make parsing fast when
it was previously impractical.  We have used the technique outlined in
this paper in other work \cite{Goodman:96b} to efficiently parse the
DOP model; in that model, the only previously known algorithm which
summed over all the possible derivations was a slow Monte Carlo
algorithm \cite{Bod:93a}.  However, by maximizing the Labelled Recall
criterion, rather than the Labelled Tree criterion, it was possible to
use a much simpler algorithm, a variation on the Labelled Recall
Algorithm.  Using this technique, along with other optimizations, we
achieved a 500 times speedup.

In future work we will show the surprising result that the last
element of Table \ref{tab:ressum}, maximizing the Bracketed Tree
criterion, equivalent to maximizing performance on Consistent Brackets
Tree (Zero Crossing Brackets) Rate in the binary branching case, is
NP-complete.  Furthermore, we will show that the two algorithms
presented, the Labelled Recall Algorithm and the Bracketed Recall
Algorithm, are both special cases of a more general algorithm, the
General Recall Algorithm.  Finally, we hope to extend this work to the
$n$-ary branching case.

\section{Acknowledgements}
I would like to acknowledge support from National Science Foundation
Grant IRI-9350192, National Science Foundation infrastructure grant
CDA 94-01024, and a National Science Foundation Graduate Student
Fellowship.  I would also like to thank Stanley Chen, Andrew Kehler,
Lillian Lee, and Stuart Shieber for helpful discussions, and comments
on earlier drafts, and the anonymous reviewers for their comments.

%%%%%%%%%%%%%%%%%%%%%%%%%%%%%%%%%%%%%%%%%%%%%%%%%%%%%%%%%%%%%%%%%%%%%%%%
%	bibliography
%%%%%%%%%%%%%%%%%%%%%%%%%%%%%%%%%%%%%%%%%%%%%%%%%%%%%%%%%%%%%%%%%%%%%%%%

%@	in submission, read in bibliography file so self-contained

\end {document}